\documentclass[letterpaper, 10 pt, conference]{ieeeconf}  

\IEEEoverridecommandlockouts                              

\usepackage{algorithmic}
\usepackage{algorithm}
\usepackage{array}
\usepackage[caption=false,font=normalsize,labelfont=sf,textfont=sf]{subfig}
\usepackage{textcomp}
\usepackage{stfloats}
\usepackage{url}
\usepackage{verbatim}
\usepackage{graphicx}
\usepackage{cite}
\usepackage{amsmath,amssymb,amsfonts}
\usepackage[utf8]{inputenc}
\usepackage{csquotes}
\usepackage{booktabs}
\usepackage[capitalize]{cleveref}
\usepackage{diagbox}
\usepackage[table,xcdraw]{xcolor}
\usepackage{threeparttable} 
\usepackage{multirow}
\usepackage{lscape}
\usepackage{xcolor}
\usepackage{float} 
\usepackage{subcaption}

\title{\LARGE \bf
An Exploratory Study on Human-Robot Interaction using Semantics-based Situational Awareness
}

\author{Tianshu Ruan$^{1}$, Aniketh Ramesh$^{1}$, Rustam Stolkin$^{1}$ and Manolis Chiou$^{2}$
\thanks{$^{1}$Tianshu Ruan, Aniketh Ramesh, and Rustam Stolkin are with Extreme Robotics Lab (ERL) and  National Center for Nuclear Robotics
(NCNR), University of Birmingham, UK.}%
\thanks{$^{2}$Manolis Chiou is the corresponding author. He is with School of Electronic Engineering and Computer Science, Queen Mary University of London, UK (m.chiou@qmul.ac.uk).}%
}

\begin{document}

\maketitle
\thispagestyle{empty}
\pagestyle{empty}

\begin{abstract}

In this paper, we investigate the impact of high-level semantics (evaluation of the environment) on Human-Robot Team (HRT) and Human-Robot Interaction (HRI) in the context of mobile robot deployments. Although semantics has been widely researched in AI, how high-level semantics can benefit the HRT paradigm is underexplored, often fuzzy, and intractable. We applied a semantics-based framework that could reveal different indicators of the environment (i.e. how much semantic information exists) in a mock-up disaster response mission. In such missions, semantics are crucial as the HRT should handle complex situations and respond quickly with correct decisions, where humans might have a high workload. Especially when human operators need to shift their attention between robots and other tasks, they will struggle to build Situational Awareness (SA) quickly. The experiment suggests that the presented semantics: 1) alleviate the perceived human operator's workload; 2) increase the operator's trust in the SA; and 3) help to reduce the reaction time in switching the Level of Autonomy (LoA) when needed. Additionally, we find that participants with higher trust in the system are encouraged by high-level semantics to use teleoperation mode more.

\end{abstract}



\section{Introduction}
\label{Introduction}
The increasing use of robots is engendering challenges and higher demands for Human-robot collaboration. HRT needs to work together and leverage complementary strengths to achieve the shared mission goal by combining perception capabilities (remote sensing, object detection) and cognition (e.g. scene understanding, decision-making, task planning). SA \cite{endsley2017toward} is one of the significant factors in bridging cognition and perception. Explainable and comprehensive SA can directly benefit robots and humans in many scopes, e.g. task planning \cite{fraune2021developing}, Variable Autonomy (VA) \cite{wagner2018overtrust}, and HRI. 

However, obtaining SA remains a challenging problem in HRT. Robots are often considered extensions of humans to perceive a remote environment using sensors through raw or low-level perception. Humans typically struggle to develop precise SA remotely, compared with the SA that we obtain when physically present in a scene \cite{mutzenich2021updating}. The problem is worse when humans shift their attention back to the robot from other tasks, e.g. when humans are out of the context/loop for a while. It takes much more effort and attention to restore the awareness of what the robot is doing and what the situation is like \cite{hergeth2016keep}, especially when humans have a limited background in the mission \cite{stanton2017human}. Moreover, SA often relies on human operators' prior experience and knowledge, which means that there is no comprehensive standard to quantify SA. 

Semantic information, which is one of the components of SA, is suggested to alleviate the above difficulties \cite{roldan2017multi}. Semantic information is widely discussed in the AI community, e.g. in computer vision. However, this literature is predominantly limited to relatively low-level semantics (e.g. image segmentation). The impact of high-level semantics, which requires more understanding and inference, has been less well-explored.

In the context of this paper, (except where specifically stated), we use the term ``semantics'' to refer to high-level semantic understanding. This can include a variety of indicators for environment understanding \cite{ruan2022taxonomy}, e.g. concepts such as risk to robots, or Signs of Human Activity (SHA).

Our previous work categorizes environmental semantics in the scope of disaster response \cite{ruan2022taxonomy} and proposes a framework \cite{ruan2025frameworksemanticsbasedsituationalawareness} that applies the above categorization and quantifies different understandings of the situation by semantic indicators. Moreover, the framework enables providing robots with an indicator of the overall semantic richness of a scene. This method translates sensor readings and detections into a situational semantics richness score, which can inform human operators if the situation requires more attention.

In this paper, we explore several high-level semantic indicators from the above framework and investigate their impact on HRT and HRI. We compare experimental robot deployments,  with and without such semantic indicators (high-level semantics), in the scope of a Search and Rescue (SAR) mission.

This paper provides new insights into HRT and HRI with respect to high-level semantics. We present three main contributions: i) identifying the usefulness of the proposed semantic richness framework \cite{ruan2025frameworksemanticsbasedsituationalawareness}; ii) identifying how high-level semantics affect HRT and HRI in the context of a SAR mission; iii) insights about patterns and relationships among HRT factors and high-level semantics, e.g. LoA preference versus trust in a VA \cite{Chiou2015TowardsTP} system.

\section{Related work}

A wide range of studies explore ways of improving SA from the perspectives of visual design, workload and stress management, human training, and LoA.

Consequently, HRI can be categorized into cognitive responses, emotional responses, behavioral responses, and physiological responses \cite{sheridan2016human}. Our work can partly be considered from a similar perspective, exploring how high-level semantics can enhance SA and benefit HRT and HRI, improving the likelihood of mission success, while reducing human workload.

Visual design significantly affects SA and workload. It covers what information to present and how to present it. In \cite{reardon2020enabling}, environmental changes were detected and displayed to the operator, by comparing differences between current and baseline LiDAR data. Such change detection information helps humans maintain SA, even in cluttered or data-dense environments. In \cite{zolotas2018head}, a rear-view display, the trajectory generated by the user’s manual input, and graphical cues were displayed on an Augmented Reality (AR) headset, reducing the need for user head rotations. This AR setup failed to increase the efficiency of the task. However, it suggested that overall task performance can be affected by many factors overlooked in the design phase, e.g. the narrower field of view of the AR headset. \cite{tabrez2022descriptive} compared two AR guidance modalities: descriptive guidance (semantic information) and prescriptive guidance (action recommendations). The authors suggested combining both modalities to enhance communication and mission efficiency or ``fluency'' in HRT. They also highlighted the importance of balancing transparency and flexibility in each guidance modality, particularly in low-certainty scenarios such as SAR missions.

\begin{figure}[htbp]
    \centering
    \includegraphics[width=7.5cm]{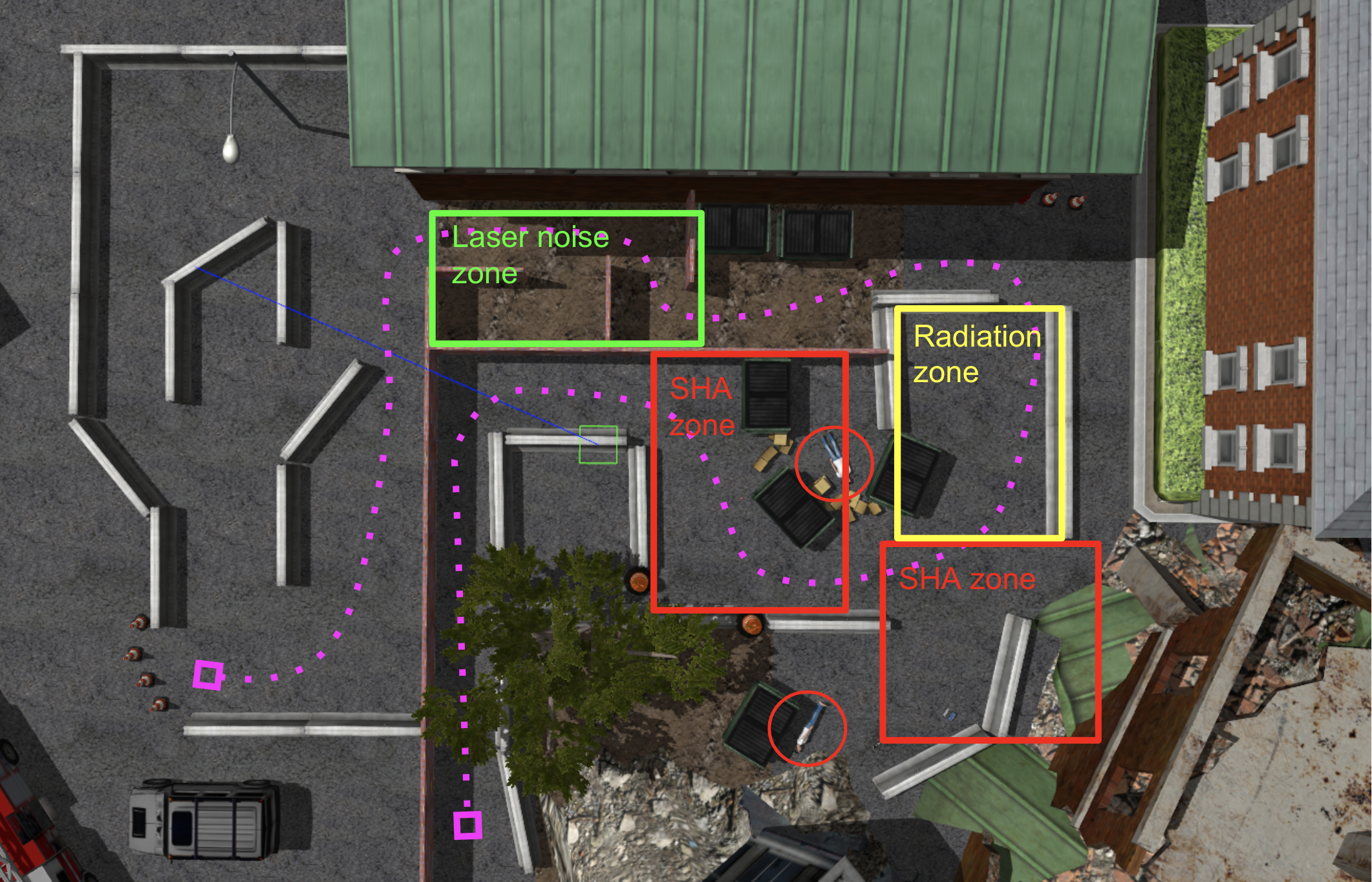}
    \caption{Test arena with semantic indicators. Colored boxes denote zones where different semantic situations will arise, resulting in different indicators being displayed to the human operator (trial participant). Red circles mark the location of the victims. The purple dotted line is the robot's pre-set autonomous navigation trajectory.}
    \label{fig: Arena semantics}
\end{figure}

\begin{figure}[htbp]
    \centering
    \includegraphics[width=7.5cm]{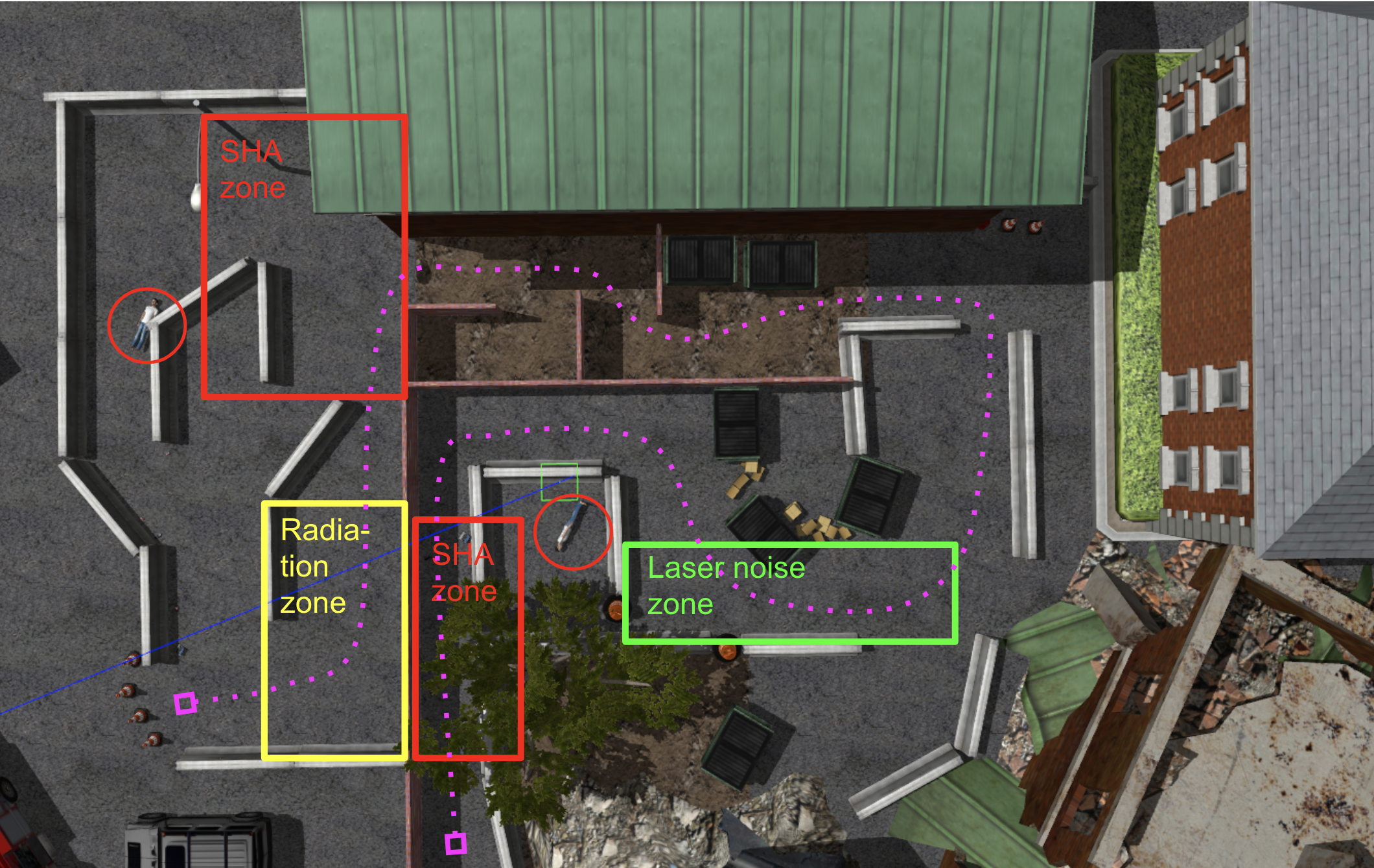}
    \caption{Test arena for trials without semantic indicators. Note that we need to use two different arena layouts for the two different trials, to minimise the confounder of learning effects between trials.}
    \label{fig: Arena no semantics}
\end{figure}

The above-mentioned studies investigated low-level semantics, but did not explore higher-level semantics. In contrast, our work explores high-level semantics, and its effects on human attention in HRT. \cite{roldan2017multi} evaluated the effects of four interfaces on human stress and workload in a simulated fire surveillance and extinguishing mission. Virtual Reality interfaces were shown to enhance SA and reduce workload compared to conventional interfaces. The authors used high-level semantics, e.g. displaying information about fire risk, during the experiments. However, they investigated and compared the effects of different interface designs for displaying such data. In contrast, our work seeks to evaluate the impact of the high-level semantic data itself, i.e. performance with versus without displaying high-level semantics.

Other work explored how human training \cite{lu2022mental} and expertise affect SA. \cite{kapellmann2016human} investigated HRT under limited SA conditions, where operators lacked a bird’s-eye view of the environment. Participants were required to navigate robots until they met up together in the experiments. The authors emphasized the importance of training. Expert operators (several hours of training) outperformed others, by minimizing the number of commands issued to the robots. This highlights the critical role of prior knowledge and training in recovering and maintaining SA.

The LoA plays an important role in affecting SA as well. \cite{gombolay2017computational} explored human factors in HRT and found that high robot autonomy could reduce human SA. This makes it difficult for humans to regain SA when taking back control of the robot. The authors recommended a mixed-initiative approach that keeps humans in the loop and balances workload and SA between humans and robots. These related studies help us identify potential confounding factors and minimize their impacts in our experiment design.

\begin{figure}[htbp]
    \centering
    \includegraphics[width=6cm]{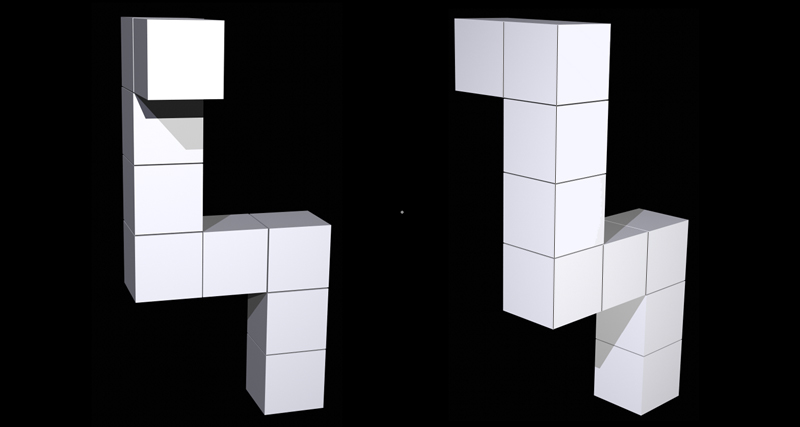}
    \caption{Parallel task example. Are these objects identical or mirror images?}
    \label{fig: Parallel task}
\end{figure}

\section{Exploratory experiment of HRI and HRT with high-level semantics}
\label{problem}

This paper explores the effects of presenting high-level semantics in HRT deployments. Specifically, we explore the performance variance of HRT, with and without the display of semantic indicators (high-level semantics). We also explore the practical utility of our previously proposed framework for real-time computing of such indicators \cite{ruan2025frameworksemanticsbasedsituationalawareness}. 

Our semantics framework \cite{ruan2025frameworksemanticsbasedsituationalawareness} quantifies the richness of low-level semantics (salient features that can be understood by a human's direct intuition) and high-level semantics (evaluation, or inference of a human's perspective) \cite{ruan2022taxonomy} in the current scene. These high-level semantic indicators can be used to help a human operator achieve SA. In our experiments, high-level semantics include: SHA, radiation levels, and navigational laser noise level. SHA is detected as a distribution of personal belongings. Radiation level is a mapping from radiation sensor readings that enables human operators to visualise radiation intensity. The laser noise refers to the robot's LiDAR. It helps robots and humans understand whether noise is becoming significant enough to disable the robot's navigation. Each of these metrics carries a normalized score between 0 and 1, helping assess potential risks or challenges in real time. 


In this exploratory study, we use a simulated SAR mission with human participants. Specifically, we simulate the first HRT deployment in a disaster response mission. The HRT needs to explore the environment and find victims while avoiding risks to robots. The robot can operate in two different LoAs: teleoperation and autonomous navigation modes. Humans can switch the LoA dynamically in a Human-Initiative paradigm \cite{manolisExperiment} and can take over teleoperative control of the robot when they feel it is necessary. Moreover, parallel cognitive tasks are assigned to human operators to simulate non-robotic emergencies that may require human attention. 


\subsection{Arena}
\label{Arena}
The experimental arena is an outdoor post-disaster environment, see \cref{fig: Arena semantics} and \cref{fig: Arena no semantics}. The robot is set to autonomously navigate along a pre-planned route, but the human operator can switch to teleoperation mode at any time, and explore other regions away from this route. Two human victims, one radiation source, and one laser noise zone are set up in the arena. Specifically, one human victim can be found by the camera (on the scheduled path, without the robot needing to deviate) by locating the personal belongings. Another victim is hidden behind obstacles, away from the scheduled path. The radiation source and laser noise zones are set to occur at pre-set regions along the planned path. The starting point and endpoint for robots are pre-defined. Note that we use a ``within groups'' experimental design, where each participant undertakes both experimental conditions (with and without semantic indicators). Therefore, we need to use different test arena layouts for each condition, to minimise learning effects as a confounder. However, the complexity of the tasks, and the types of semantics encountered, are kept very similar to ensure meaningful comparability.

\subsection{Mission and Tasks}
We assume that the HRT needs to conduct its first deployment to investigate the site in a SAR mission. The participant can switch back and forth between the teleoperation and autonomy. In autonomy mode, the robot will try to follow the pre-set route. Three tasks are assigned to the HRT: \textbf{A)} finding victims, \textbf{B)} navigating the robot, and \textbf{C)} parallel task. Task \textbf{A} requires the HRT to find as many victims as they can in the arena. Task \textbf{B} requires the HRT to navigate the robots from a pre-set starting point to the endpoint within set time limits. Task \textbf{C} is a non-robotics task, used to simulate additional types of cognitive loading. This additional cognitive loading encourages participants to make use of the LoA switching functionality, and helps elucidate situations where the additional semantic information becomes useful. Participants are instructed to do their best on all tasks (equal importance). They conduct two trials comparing two conditions: semantics presented \cref{fig: Arena semantics}; semantics not presented \cref{fig: Arena no semantics}. To minimize the learning effects, the sequence of two trials is counterbalanced among participants. Moreover, victims' locations are different in the two trials, and starting/end points are swapped (clockwise/anti-clockwise path) between participants.

Participants have five minutes for each trial. If they fail to bring the robot back to the endpoint within the time limit, the mission will be marked as a failure. Participants are not informed of the number of victims. They need to maximize their utilization of the time. Hence, a time-to-completion metric is not meaningful. Instead, we examine success rates on each task.

The parallel task takes place continuously (an endless task) during the mission. Participants need to identify whether a pair of 3D objects are identical, or are mirror images, see \cref{fig: Parallel task}. This task has been validated in \cite{ganis2015new}. Standardized training \cite{chiou2019learning} was provided on this task before the experiments to ensure all participants were aware of the effort required to manage this alongside the two robotics tasks.

Participants' performances on this task (accuracy, completion, and completion speed), along with the robot-related tasks, are included in the overall evaluation. In addition, we recorded the baseline (accuracy of ten questions) of each participant before the experiment and compared the baseline to the parallel task accuracy during the experiments. From this comparison, we can evaluate the impact of high-level semantics on participants’ parallel task execution quality. 


\subsection{Experimental design}
A total of 16 participants were recruited from university staff and students aged 22-50, see \cref{tab: background}. Most had a driver's license or experience with 3D video games, which correlate with controlling remote vehicles \cite{chiou2015towards}. Participants undertook standardised training to understand the system functionalities and control interface needed for the mission, including navigation in teleoperation and autonomy modes, robot's SLAM map, and collision avoidance. Participants were introduced to high-level semantics, including SHA, radiation, and laser noise. When a semantic indicator score increases, the system emits a warning sound to attract human attention. Participants were trained on how to react to situations, e.g. switch LoA to teleoperation when needed. Training is designed to avoid very large individual performance differences, especially if participants become lost in the arena. 

\begin{table}[h!]
\centering
\begin{threeparttable}
\caption{Participants' gender and background}
\begin{tabular}{cccc}
\toprule
\textbf{Gender} & \textbf{Participants} & \textbf{Background} & \textbf{Participants}  \\ 
\midrule
Male   & 8/\textcolor{red}{7}*  & Driver only     & 5/\textcolor{red}{5}    \\ 
Female & 8/\textcolor{red}{7}  & Gamer only      & 2/\textcolor{red}{2}    \\ 
       &                       & Gamer + Driver  & 7/\textcolor{red}{6}    \\ 
       &                       & Neither both    & 2/\textcolor{red}{1}    \\ 
\bottomrule

\end{tabular}

\begin{tablenotes}
    \footnotesize
    \item {* Red digits refer to the number of participants after data cleaning, which are involved in the future analysis.}
    \end{tablenotes}
    \label{tab: background}
\end{threeparttable}
\end{table}

To minimize the effects of potential confounding factors, we surveyed participants' background information in \cref{tab: background} and their feedback on arena and mission tasks. We removed two participants' records from the database. One had a pre-existing bias toward the autonomy system, with no use of autonomy mode. She explained she had ``zero trust in any robot system''. Another one failed to follow the instructions given, which led to incomplete data. The red digit in \cref{tab: background} refers to the number of participants after removal.

\begin{figure}[htbp]
    \centering
    \includegraphics[width=8cm]{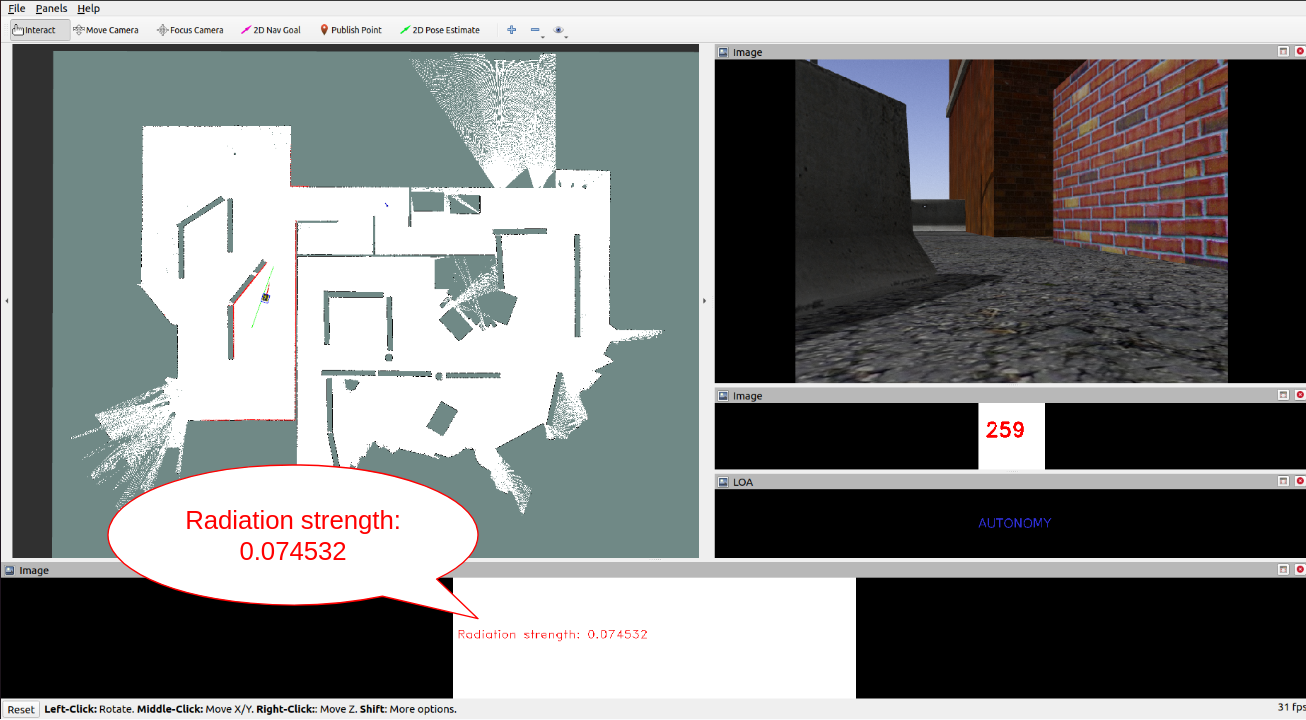}
    \caption{GUI for trials without semantic indicators: participants obtain SA from a pre-scanned LiDAR map, camera feed, real-time localisation, radiation readings, current robot LoA, and countdown of remaining mission time. The bubble is used for clearly presenting the semantic indicator scores. It does not show up in the GUI during the experiments.}
    \label{fig: UI no semantics}
\end{figure}

\begin{figure}[htbp]
    \centering
    \includegraphics[width=8cm]{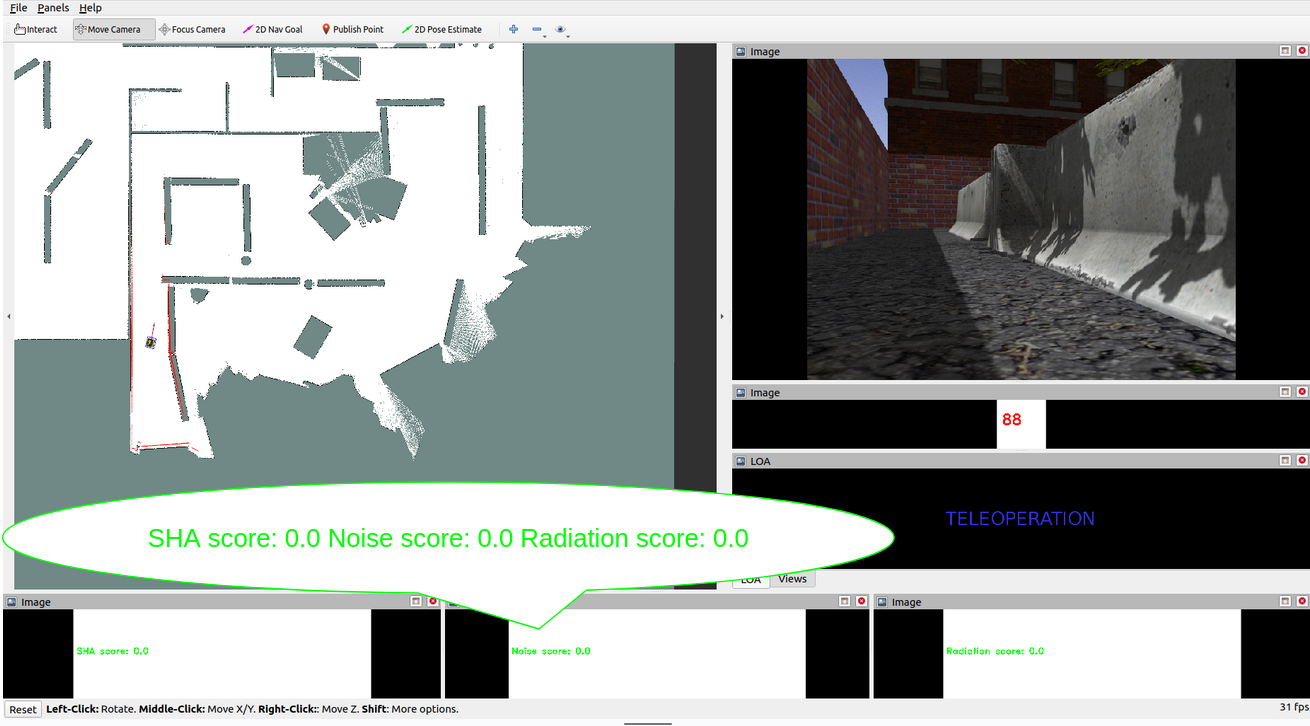}
    \caption{GUI for trials with semantic indicators: three semantic indicator scores are displayed on the interface screen, in addition to the mapping and video data.}
    \label{fig: UI semantics}
\end{figure}

The GUIs of the two experimental conditions are presented in \cref{fig: UI no semantics} and \cref{fig: UI semantics}.

\begin{figure*}
\centering
\subfloat[Number of victims found]{
		\includegraphics[scale=0.32]{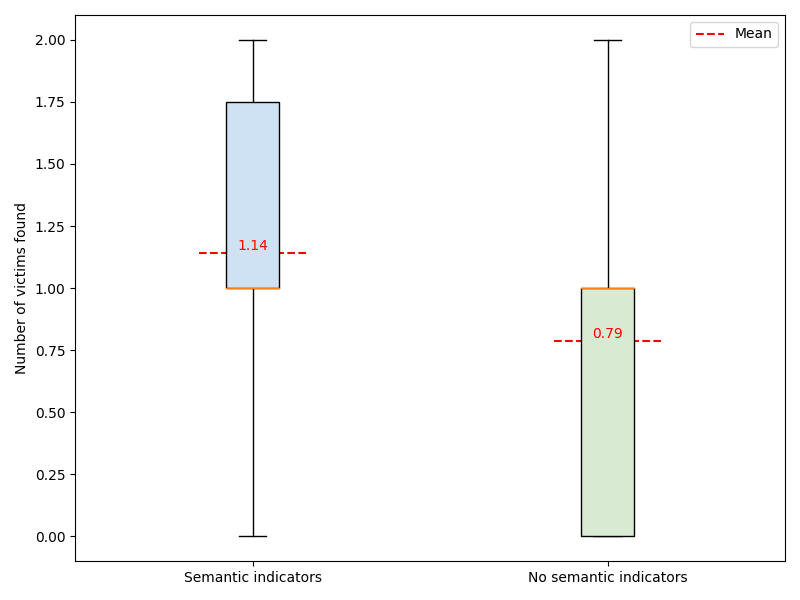}}
\subfloat[Parallel task accuracy]{
		\includegraphics[scale=0.32]{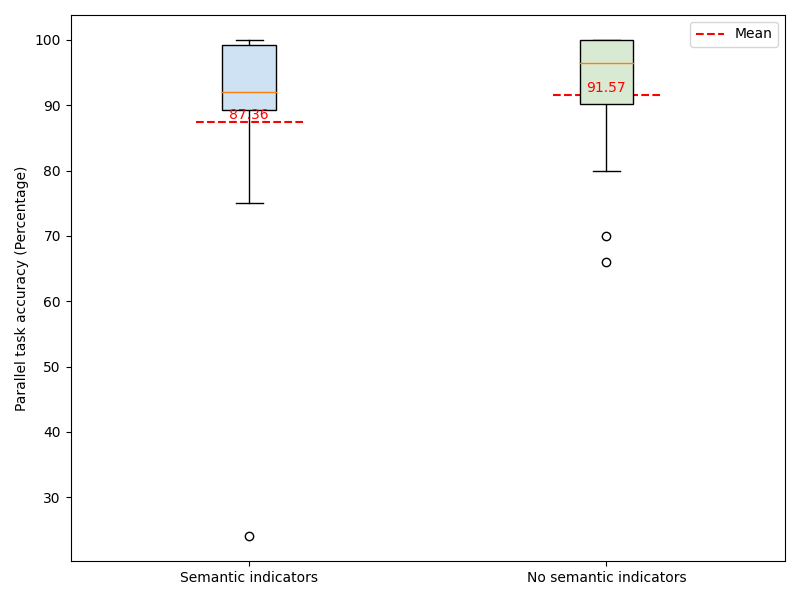}}
\\
\subfloat[Parallel task completion speed]{
		\includegraphics[scale=0.32]{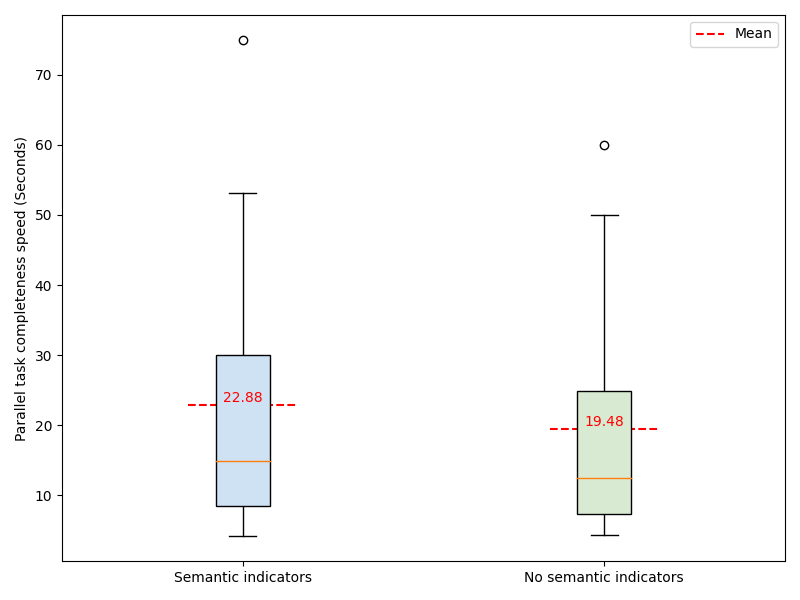}}
\subfloat[Teleoperation ratio]{
		\includegraphics[scale=0.32]{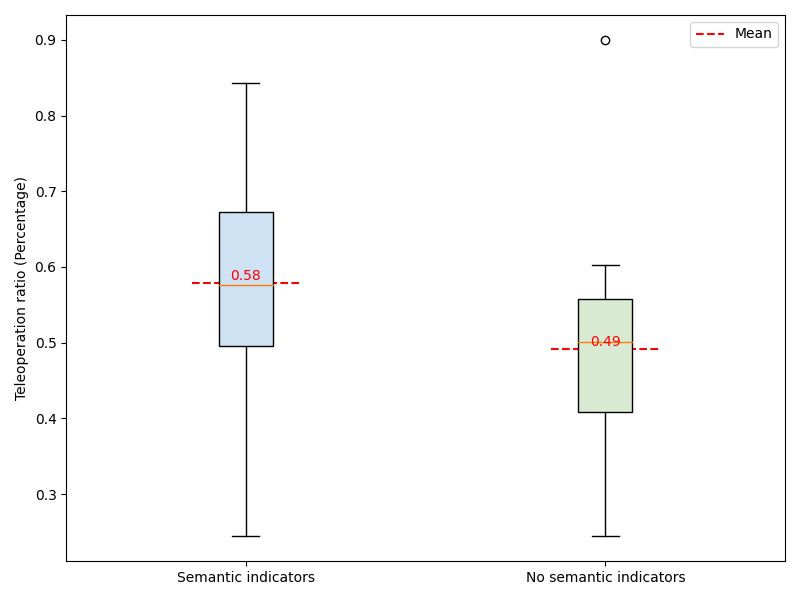}}
\\
\subfloat[Number of LoA switches]{
		\includegraphics[scale=0.32]{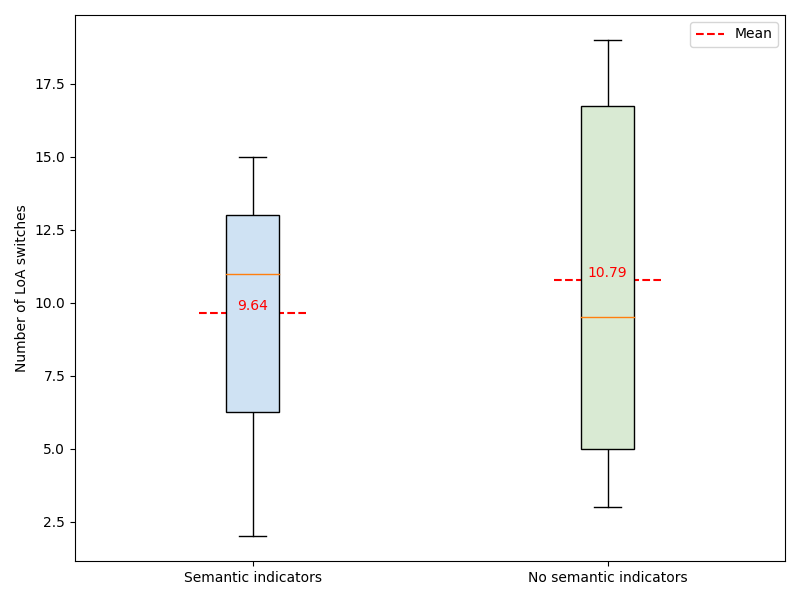}}
\subfloat[Response time to switch LoA]{
		\includegraphics[scale=0.32]{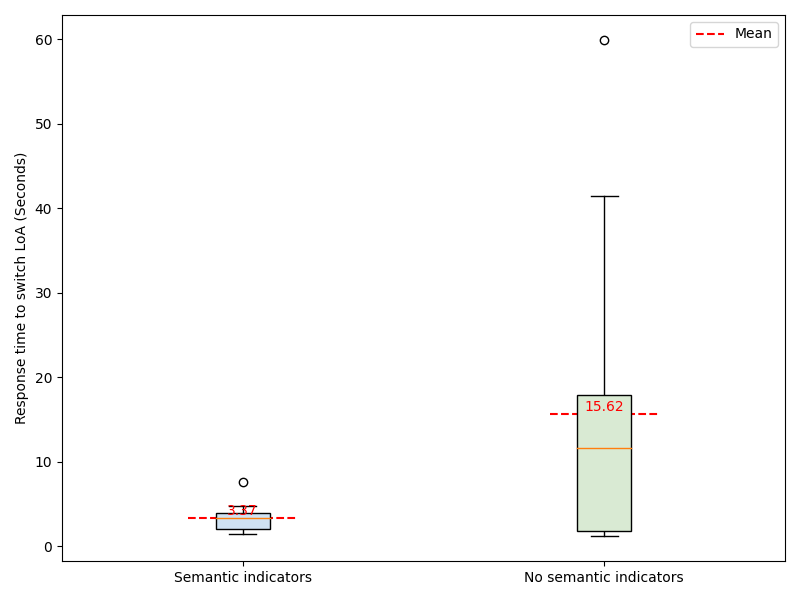}}
\\
\subfloat[NASA TLX]{
		\includegraphics[scale=0.32]{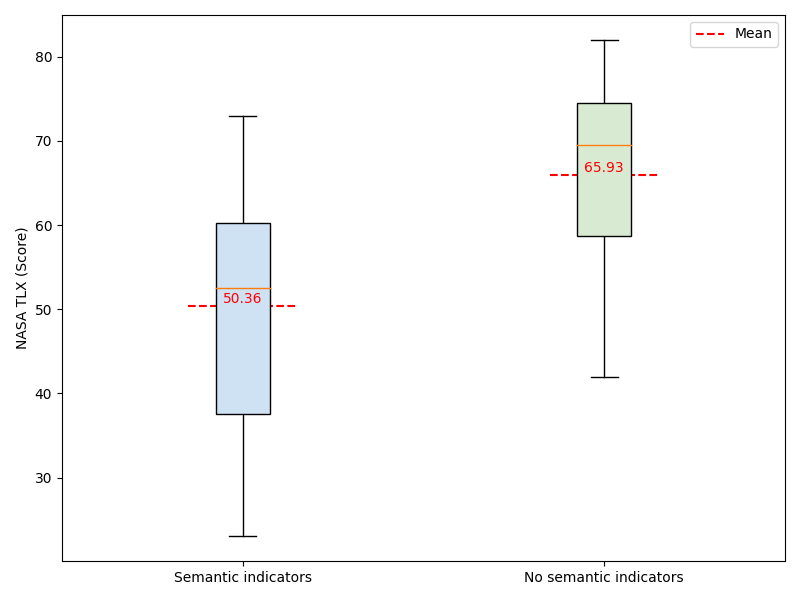}}
\subfloat[Situational trust scale]{
		\includegraphics[scale=0.32]{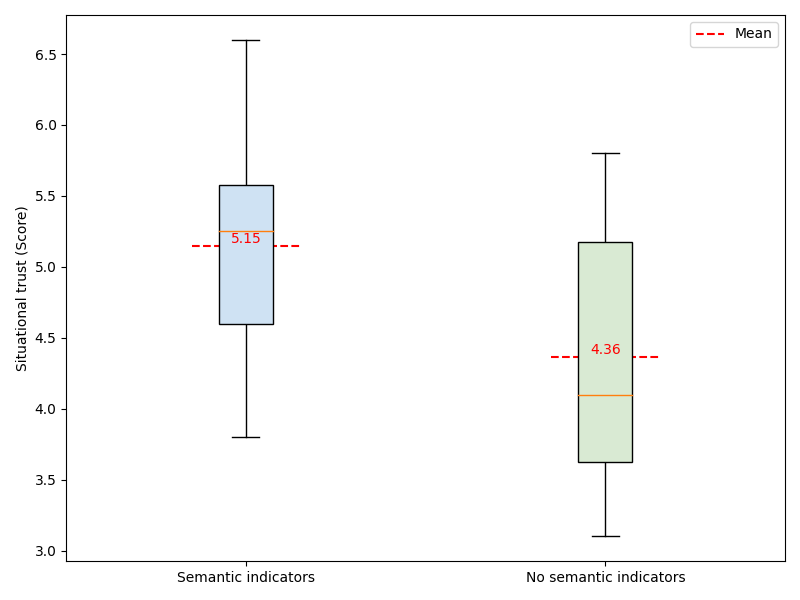}}
\caption{Metrics comparison between trials with semantic indicators and trials without semantic indicators. The mean value of each set is marked with a red dashed line with a number, and the orange line refers to the median value.}
\label{Comparision}
\end{figure*}

The above training ensured that participants understood the system features and operation modes, creating a foundation for effective decision-making and navigation. Moreover, training gave participants an understanding of the workloads to help them make decisions on how to utilise the VA and semantic indicator functionalities. To minimize the effects of various confounders, some additional information and tips were given to participants:

\begin{itemize} 

\item Because of blind spots from the LiDAR, the autonomy mode might fail in some places. Participants may need to switch to teleoperation mode.
\item The time limitation is tight for most operators. Participants should carefully manage the time and areas to explore. For the same reason, try to follow the scheduled (shortest) path most of the time, and only deviate when participants feel it is necessary.
\item The three semantic indicators will not appear at the same time.
\item When a situation changes (e.g. SHA appear), participants are recommended to switch to teleoperation mode.
\end{itemize} 

We used a Jackal mobile robot with a Velodyne 3D LiDAR, a Realsense D435i camera, and a Hamamatsu gamma ray sensor in the simulated arena. The robot navigated in the arena using customized AMCL and movebase ROS packages, with the same top speed limit in autonomy and teleoperation modes. Participants could switch LoA at will.

\section{Results}
In the experiments, we collected the following data:
\begin{itemize} 

\item {Objective measurements:} number of victims found, parallel task accuracy, parallel task completed, parallel task completion speed, parallel task accuracy degradation, reaction time to situation changes, number of LoA switches, percentage of teleoperation use (teleoperation ratio), navigation task completion.

\item {Subjective measurements:} NASA TLX \cite{TLX}, situational trust scale for automated driving \cite{reardon2020enabling}, and trust in the system.
\end{itemize} 

Specifically, parallel task completion is the number of the completed questions in the parallel task; parallel task completion speed refers to the average time taken in seconds for each parallel task question; parallel task accuracy degradation is the negative accuracy variance compared with the baseline; navigation task completion is the mark to label if the participant successfully navigate the robot to the endpoint within time limits (``0'' is failure, ``1'' is success); NASA TLX applies a questionnaire to evaluate human workload; situational trust scale for automated driving is derived from a questionnaire which evaluates trust in the obtained SA; trust in the system is a metric from the situational trust scale, rating from 1 (low) to 7 (high).

\subsection{Significance analysis}

\begin{table}[h]
\centering
\caption{Significance analysis using Wilcoxon signed-rank test (\textit{n} = 14)}
\label{tab:significance}
\begin{tabular}{lccc}
\toprule
\textbf{Metric} & \textbf{$W$} & \textbf{$P$} & \textbf{$r$} \\
\midrule
Number of victims found & 0.0 & 0.102 & 0.436 \\
Parallel task accuracy   & 17.0 & 0.514 & 0.174 \\
Parallel task accuracy degradation  & 17.0 & 0.515 & 0.174 \\
Parallel task completion & 42.0 & 0.542 & 0.163 \\
Parallel task completion speed & 31.0 & 0.194 & 0.347 \\
Number of LoA switches & 33.0 & 0.241 & 0.313 \\
Teleoperation ratio & 30.0 & 0.173 & 0.365 \\
Trust in the system & 10.5 & 0.080 & 0.468 \\
\textbf{Response time to switch LoA} & \textbf{16.0} & \textbf{0.040} & \textbf{0.570} \\
\textbf{NASA TLX} & \textbf{11.0} & \textbf{0.007} & \textbf{0.724} \\
\textbf{Situational trust scale} & \textbf{11.5} & \textbf{0.031} & \textbf{0.577} \\
Navigation task completion & 15.0 & 0.317 & 0.267 \\
\bottomrule
\end{tabular}
\end{table}

Given the number of participants and no assumption about the data, we conduct a significance analysis using the Wilcoxon signed-rank test, a non-parametric method. We calculate the W-statistic value $W$, the P-value $P$, and the effect size $r$ for each measurement. We consider a result statistically significant when $P$ is $<0.05$, and a meaningful and strong practical difference when $r$ is $>0.5$ according to Cohen's standard \cite{cohen2013statistical}. We did not use Bonferroni or related p-value corrections as the study is exploratory.  

We observed statistically significant differences in the reaction time to situation changes, the NASA TLX, and the situational trust scale. \cref{tab:significance} shows that response time to switch LoA is significantly lower and shows a strong practical difference ($W$ = 16.0, $P$ = 0.040, $r$ = 0.570) with semantic indicators. It means high-level semantics significantly help to lower the reaction time to situation changes (e.g. a new detection of personal belongings). NASA TLX is significantly lower and shows a strong practical difference ($W$ = 11.0, $P$ = 0.007, $r$ = 0.724). It refers that semantic indicators lower the perceived human workload. Moreover, the situational trust scale is significantly higher and shows a strong practical difference ($W$ = 11.5, $P$ = 0.031, $r$ = 0.577) with the semantic indicators. It means participants have more trust in their SA with the assistance of semantic indicators.

No significant differences are observed in the metrics: parallel task accuracy, parallel task accuracy degradation, parallel task completion, parallel task completion speed, number of LoA switches, and navigation task completion. However, for the number of victims found and the trust in the system, they show a trend for finding more victims with the semantic indicators.

\subsection{Correlation analysis}
\begin{figure*}[htbp]
    \centering
    \includegraphics[width=19cm]{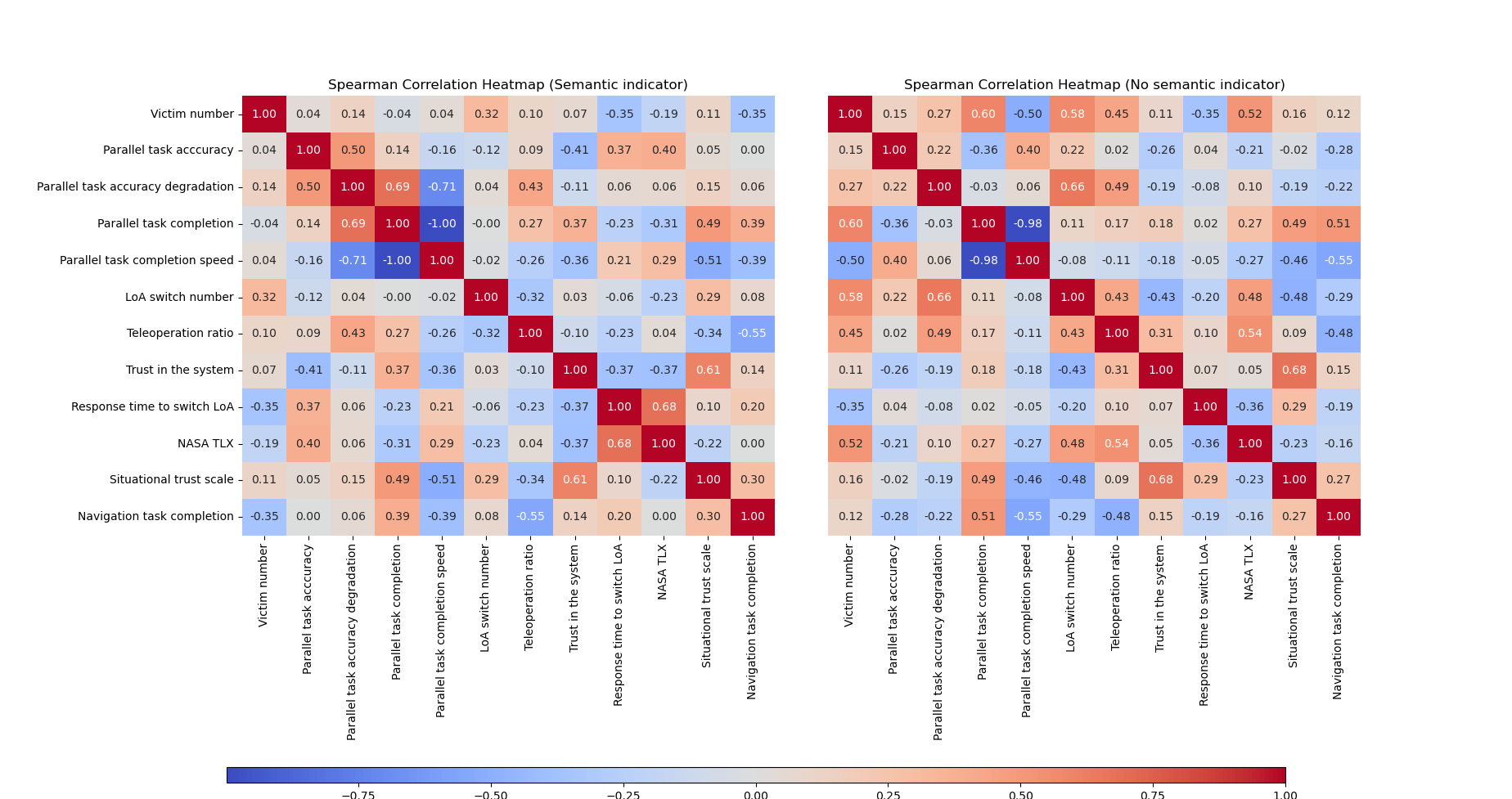}
    \caption{Spearman’s $\rho$ correlation heatmaps of experimental trials with and without semantics.}
    \label{fig: heatmap}
\end{figure*}

We employ Spearman’s correlation analysis to find the hidden relationship/patterns among these factors. The correlation heat map of each measurement is presented in \cref{fig: heatmap}. With the comparison of the two heatmaps, we observe some interesting findings with strong correlations. They are: 
\begin{itemize} 

\item {} The parallel task accuracy has a positive mild correlation with the NASA TLX ($\rho$ = 0.40) in the semantics trial, i.e. when the parallel task accuracy is high, the cognitive workload is high. However, this shows a negative correlation ($\rho$ = -0.21) in the non-semantics condition. I.e., when the parallel task accuracy is high, the cognitive workload appears slightly decreased.

\item {} The parallel task completion speed is significantly negatively correlated with parallel task accuracy degradation ($\rho$ = -0.71) in the semantics trial. This suggests that, when the parallel task completion speed is high, the parallel task accuracy is higher than baseline. However, this becomes non-correlated ($\rho$ = -0.06) in the non-semantics trial.

\item {} The trust in the system has a mild negative correlation with the NASA TLX ($\rho$ = -0.37) in the semantics trial. I.e. when trust in the system is high, cognitive workload is low. However, this shows non-correlated in the non-semantics trial ($\rho$ = 0.05).

\item {} Trust in the system has a mild negative correlation with the LoA switch number ($\rho$ = -0.43) in the non-semantics trial. This suggests that, when trust in the system is high, the number of LoA switches is low. However, this shows little correlation ($\rho$ = 0.03) in the semantics trial.

\end{itemize}

\section{Discussion}

The significance analysis indicates that high-level semantics have impacts on response time to switch LoA, NASA TLX, and situational trust scale. Specifically, with the help of the semantic indicators, human operators could quickly switch LoA ($W$ = 16.0, $P$ = 0.04, $r$ = 0.570) when the situation changes (e.g. new personal belongings detected). And, NASA TLX significantly decreases ($W$ = 11.0, $P$ = 0.007, $r$ = 0.724), indicating that semantic indicators help to reduce overall cognitive load. Moreover, the situational trust scale significantly increases ($W$ = 11.5, $P$ = 0.031, $r$ = 0.577), which means human operators have more trust in their SA when assisted by semantic indicators.

Besides the above significant changes, the analysis shows the trend that high-level semantics might have positive effects on the number of victims found ($W$ = 0, $P$ = 0.102, $r$ = 0.436) and trust in the system ($W$ = 10.5, $P$ = 0.080, $r$ = 0.468). These trends have the potential to become significant impacts if we involve more participants in the experiment. In the end, no significant differences were detected in rest indicators (parallel task accuracy, completion rate, navigation completion, etc.), suggesting that simply adding high-level semantics does not improve these performances. They may still be affected by task design and other factors in the experiment.

In summary, the significance analysis suggests that using high-level semantic indicators, based on our framework can significantly benefit HRT with respect to: response time to switch LoA; workload of the human; and the operator's trust in their SA. High-level semantic indicators shows a trend that they help to increase the number of victims found and human's trust in the robotic system.

The correlation analysis reveals some additional interesting findings:


\begin{itemize}
\item The correlation between parallel task accuracy and the NASA TLX suggests that the high-level semantic indicators may be providing useful additional information, but this requires additional cognitive workload. If participants seek to improve the parallel task performance by exploiting the semantic indicator displays, this might cause additional workload ($\rho$ = 0.40 suggests a moderate positive monotonic relationship). Hence, even though high-level semantics helps alleviate cognitive workload, this may cause the human to exert additional effort to improve the parallel task accuracy, beyond what might otherwise have been possible without the semantic indicators. A similar finding is claimed by \cite{endsley2018automation} that ``subjects may have been able to distribute their excess attention to other displays and tasks", if the automation improves SA by reducing excessive workload.
\item Correlation between trust in the system and NASA TLX (with-semantics trial) suggests that, when participants have a lower workload, they may trust the system more. However, the non-semantics trial shows no such correlation. By providing the high-level semantics, human participants could strengthen their trust in the system from the saved ``bandwidth" of cognitive load \cite{endsley2017here}.
\item Trust shows negative correlation with the number of LoA switches (non-semantic trials), but little correlation in with-semantics trials. Some researchers \cite{wangmodelling} claim that ``higher trust usually results in an operator accepting the automation’s decisions." Human operators take more control when trust goes down. However, we believe that there are more complex reasons hidden behind. Participants with low system trust make many LoA switches (perhaps ``noisy decision making'' \cite{hilbert2012toward}) when semantic indicators are unavailable. Noisy decision making could easily happen when humans are in a high-uncertainty situation (poor SA). By introducing semantic indicators (lowering uncertainty and improving SA), human operators might avoid dropping into that high-uncertainty situation.
\end{itemize}

\begin{figure}[htbp]
    \centering
    \includegraphics[width=7.2cm]{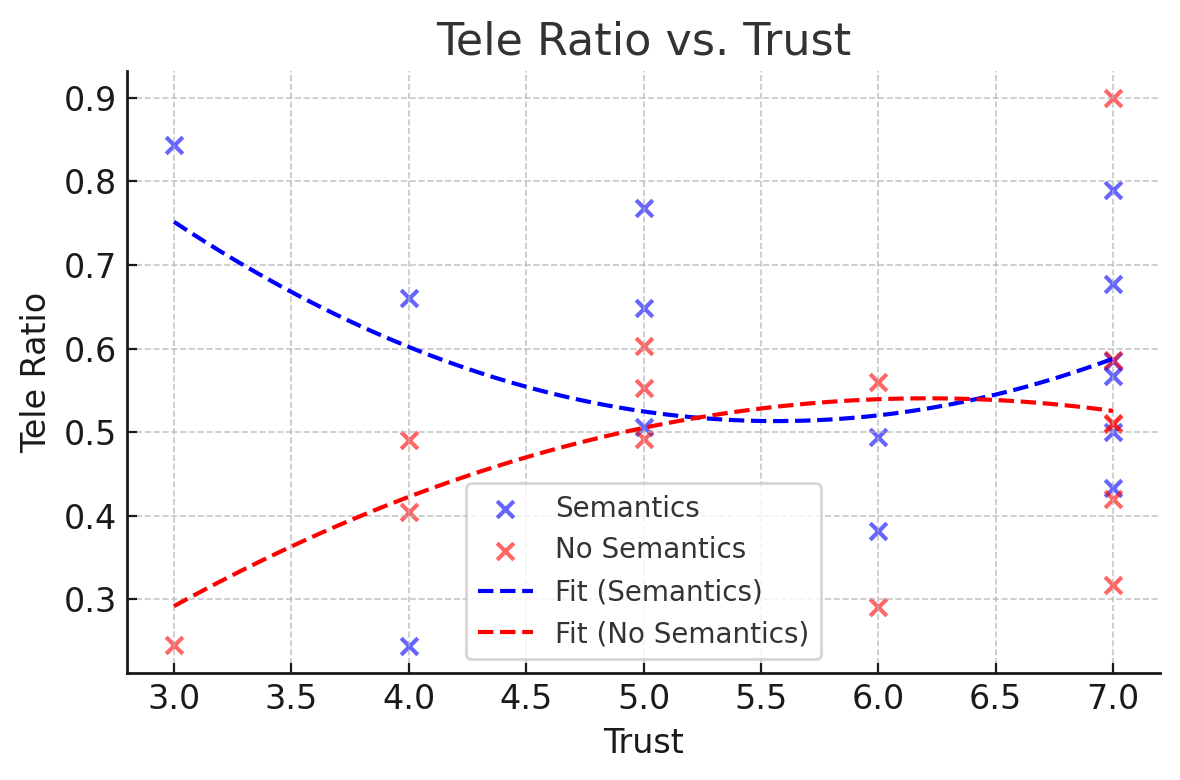}
    \caption{Teleoperation ratio vs. trust: red dash curve is no-semantics condition. Blue dash curve is with-semantics condition.}
    \label{fig: trust}
\end{figure}

The box plots of \cref{Comparision} suggest a correlation between high trust in the robotic system, and large amounts of driving the robot in teleoperation mode. Meanwhile, the correlation between trust and the number of LoA switches in different experiments suggests an interesting relationship, prompting us to explore these relationships further. In \cref{fig: trust}, we see that participants with low trust and no semantics strongly favour autonomy mode. In contrast, with semantic indicators, participants with low trust strongly favour teleoperation. This suggests that: \textbf{i)} the semantic indicator displays comprise a large part of participants' SA, i.e. displaying these semantic indicators causes participants to feel they have a much better understanding of the remote scene surrounding the robot; \textbf{ii)} when participants trust in their SA, they feel much more confident about using teleoperation mode; \textbf{iii)} they prefer teleoperation mode (provided they have sufficient SA) when they have low trust in the robotic ``system'' i.e. in the autonomous agent. Furthermore, we can see that, at high levels of trust, participants converge on equal use of autonomy and teleoperation, regardless of the amount of SA available to them.

\section{Limitations}
To minimize learning effects as a confounder in our ``within-groups'' experimental design, we used (slightly) different arena layouts between the two trials. This might have introduced a difference in overall difficulty between the two trials, although we tried to design the two arenas to have similar difficulties. To check this, we asked participants which arena they thought was more difficult. Five participants thought both arenas were equally easy. Three thought the with-semantics arena was more difficult. Six thought the without-semantics arena was more difficult. While not identical, these results suggest similar difficulties. Note also that the without-semantics condition might make the without-semantics arena seem more difficult, further reassuring that the arenas had similar overall difficulty.


Some participants waited for the sound warning (3 seconds) to end and before they switched LoA. This unnecessarily increased the human idle time. It indicates multimodal notification, such as vibrations on the joystick, could be a better option for delivering this semantics.

\section{Conclusion}

In this paper, we validated the usefulness of our proposed framework \cite{ruan2025frameworksemanticsbasedsituationalawareness},  and investigated the effects of high-level semantic indicators in HRT and HRI. Moreover, we explored VA HRT patterns when high-level semantics are involved. Our experiments indicate that displaying high-level semantic indicators can help humans: decrease reaction time when switching LoA, reduce cognitive workload, and increase trust in their SA.




\bibliographystyle{IEEEtran}
\bibliography{cit}

\end{document}